\documentclass[pdflatex,sn-mathphys-num]{sn-jnl}% Math and Physical Sciences Numbered Reference Style
\usepackage{graphicx}%
\usepackage{multirow}%
\usepackage{amsmath,amssymb,amsfonts}%
\usepackage{amsthm}%
\usepackage{mathrsfs}%
\usepackage[title]{appendix}%
\usepackage{xcolor}%
\usepackage{textcomp}%
\usepackage{manyfoot}%
\usepackage{booktabs}%
\usepackage{algorithm}%
\usepackage{algorithmicx}%
\usepackage{algpseudocode}%
\usepackage{listings}%
\usepackage{tikz}

\theoremstyle{thmstyleone}%
\theoremstyle{thmstyletwo}%

\theoremstyle{thmstylethree}%

\setcitestyle{super}

\begin{document}

\title[Article Title]{The Fourier Spectral Transformer Networks For Efficient and Generalizable Nonlinear PDEs Prediction}


\author*[1]{\fnm{Beibei} \sur{Li}}\email{blb0607@gmail.com}

\affil*[1]{\orgname{Deep Space Exploration Lab}, \country{China}}


\abstract{
In this work we propose a unified Fourier Spectral Transformer network that integrates the strengths of classical spectral methods and attention based neural architectures. By transforming the original PDEs into spectral ordinary differential equations, we use high precision numerical solvers to generate training data and use a Transformer network to model the  evolution of the spectral coefficients. We demonstrate the effectiveness of our approach on the two dimensional incompressible Navier-Stokes equations and the one dimensional Burgers’ equation. The results show that our spectral Transformer can achieve highly accurate long term predictions even with limited training data, better than traditional numerical methods and machine learning methods in forecasting future flow dynamics. The proposed framework generalizes well to unseen data, bringing a promising paradigm for real time prediction and control of complex dynamical systems.}




\maketitle

\section{Introduction}\label{sec1}

The partial differential equations solving is a cornerstone of computational science and engineering. Traditional methods can be computationally expensive, especially when dealing with high-dimensional, nonlinear, and complex systems. 

In recent years, machine learning has emerged as a powerful tool to address these challenges, particularly through the integration of data-driven models with physics-based constraints. The key development in this area is Physics-Informed Neural Networks, which incorporate PDE residuals and boundary and initial condition penalties directly into the loss function of a neural network, unifying forward and inverse problem solving \cite{Raissi2019PINN}. The comprehensive survey of physics‐informed machine learning elucidates the connections between PINNs, neural operators, neural differential equations, and outlines future challenges \cite{Karniadakis2021PR}. PINNs in Fluid Mechanics \cite{Cai2022Fluid} reviewed the application of PINNs in fluid mechanics, summarizing their performance on Navier–Stokes, Burgers’, and other flow problems, and discussing practical aspects like loss weighting and adaptive sampling. The Adaptive Activation in PINNs \cite{Jagtap2020Adaptive} introduced adaptive activation functions in PINNs, showing that dynamically learned nonlinearities can accelerate convergence and improve accuracy for stiff PDEs. Hidden Fluid Mechanics \cite{Raissi2020HFM} showed how PINNs can infer hidden fluid dynamics recovering velocity and pressure fields purely from sparse flow visualizations thus enabling non‐intrusive flow diagnostics from image data. 

The Neural Operators have emerged as another significant advancement. These models, such as the Fourier Neural Operator, learn the solution operators for parametric PDEs by performing convolution in the Fourier domain \cite{Li2021FNO}. The introduction of Dealiasing FNO further improves numerical precision by correcting spectral aliasing errors, a key limitation in high-frequency domains \cite{Tran2023Dealiasing}. DeepONet, a framework for learning nonlinear operators between function spaces, also provides universal approximation capabilities for parametric PDEs \cite{Lu2021DeepONet}. The Graph Kernel Network\cite{li2020gkn} is an early operator‐learning framework that laid the groundwork for FNO. Kovachki \emph{et al.}\cite{kovachki2021lo} provided a unified theoretical analysis of neural operators including FNO, DeepONet, and GKN highlighting their approximation properties. Gupta \emph{et al.}\cite{gupta2022mw} proposed a multiwavelet‐based operator learning approach, extending the spectral operator paradigm to wavelet bases. Pathak \emph{et al.}\cite{pathak2022fourcastnet} applied FNO in FourCastNet for high-resolution global weather forecasting, demonstrating state-of-the-art predictive performance.
Li \emph{et al.}\cite{li2022pino} combined FNO with physics constraints to form the Physics‐Informed Neural Operator. Lu \emph{et al.}\cite{lu2022benchmark} conducted a comprehensive benchmark of deep operator networks, comparing FNO against alternatives on unstructured PDE problems and analyzing architectural trade‐offs.

There are other important contributions including Sinusoidal Representation Networks which use periodic sine activation functions to significantly improve the ability of neural networks to model high frequency signals and solve differential equations improving the approximation of implicit fields and high frequency components \cite{Sitzmann2020SIREN, tancik2020fourier, ramasinghe2023inr, kloock2023survey}. The Transformers for PDEs explore the use of attention-based architectures to model non-local interactions in PDE solutions, offering a promising method for dealing with complex dependencies between variables \cite{Poli2022Transformers}. The integration of classical methods with neural networks has led to the development of Learned Iterative Solvers, which combine neural networks with traditional iterative solvers to accelerate convergence while maintaining physical consistency \cite{Ke2023LIS}.

The machine learning methods have not yet surpassed traditional numerical approaches in terms of accuracy. In particular, convergence and stability analyses remain limited compared to those available for conventional numerical methods, and predictive accuracy outside the training domain is often insufficient. We introduce a Fourier Spectral Transformer network. The results demonstrate that the spectral Fourier Transformer has the potential for accurate long term prediction after training. The neural network is capable of producing highly accurate predictions even when trained on relatively limited data. Its ability to generalize from sequences enables robust forecasting of future velocity fields with minimal computational cost a significant advantage over traditional methods. The network effectively captures the flow dynamics throughout both the training and prediction intervals, proving its potential as a powerful tool for solving complex fluid dynamics problems.
%However, machine learning methods have not yet surpassed traditional numerical approaches in terms of accuracy. In particular, convergence and stability analyses remain limited compared to those available for conventional numerical methods, and predictive accuracy outside the training domain is often insufficient. To address these challenges, we introduce a Fourier Spectral Transformer network. Our results demonstrate that the spectral Fourier Transformer has the potential for accurate long-term prediction after training, in some cases outperforming standard numerical methods. Notably, the neural network is capable of producing highly accurate predictions even when trained on relatively limited data. Its ability to generalize from temporal sequences enables robust forecasting of future velocity fields with minimal computational cost—a significant advantage over traditional methods. Furthermore, the network effectively captures the flow dynamics throughout both the training and prediction intervals, further underscoring its promise as a powerful tool for solving complex fluid dynamics problems.

\section{The Fourier Spectral Network Structure}
We introduce a unified modeling and forecasting framework for partial differential equations, combining spectral methods with neural network architectures.
\paragraph{PDE Formulation}
Consider a general evolution PDE of the form:
\begin{equation}
    \frac{\partial \mathbf {u}(\mathbf{x}, t)}{\partial t} = \mathcal{G}\bigl(\mathbf{u}(\mathbf{x}, t), \nabla \mathbf{u}(\mathbf{x}, t), \dots\bigr),
\end{equation}
where $\mathbf{u}(\mathbf{x},t)$ is the field variable defined on the spatial domain $\mathbf{x}$, and $\mathcal{G}$ is the evolution operator e.g., advection, diffusion, source terms, which may include linear or nonlinear contributions.

\paragraph{The Spectral Transformation: Mapping PDE to ODE}
Expand $\mathbf{u}(\mathbf{x},t)$ in a Fourier spectral basis to derive a system of ordinary differential equations in spectral space:
\begin{equation}
    \mathbf{u}(\mathbf{x}, t) = \sum_{k} \hat{\mathbf{u}}_k(t) \phi_k(\mathbf{x}),
    \qquad
    \frac{d\hat{\mathbf{u}}_k}{dt} = \langle \phi_k, \mathcal{G}(\mathbf{u},\nabla \mathbf{u},\dots) \rangle =: \mathcal{F}_k(\{\hat{\mathbf{u}}_j\}, t),
\end{equation}
where $\phi_k(\mathbf{x})$ are the basis functions, $\langle\cdot,\cdot\rangle$ denotes the inner product, and $\mathcal{F}_k$ is the spectral evolution operator for mode $k$.

\begin{figure}
        \includegraphics[width=\linewidth]{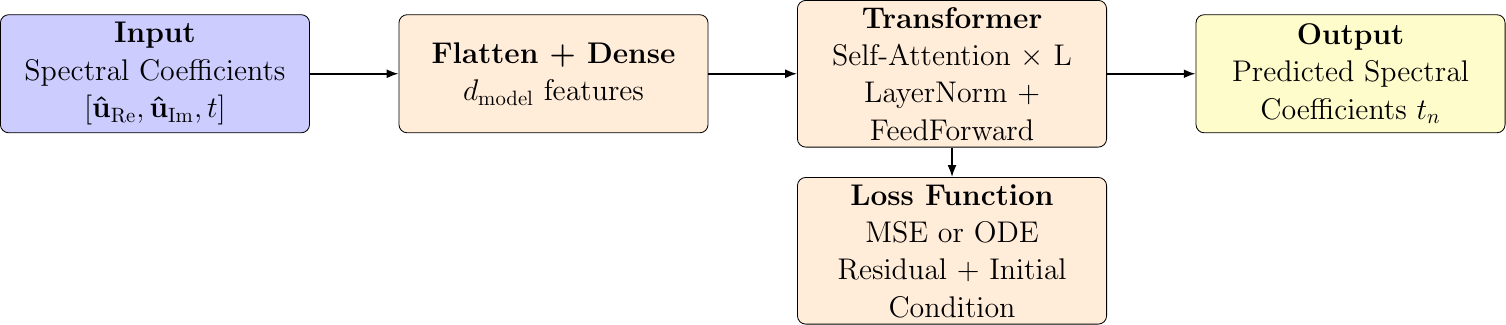}
    \caption{The Fourier Spectral Transformer Structure}
    \label{fig:tran}
\end{figure}

Building on this mapping, we proceed with following steps
\begin{enumerate}
    \item \textbf{Numerical Integration and Data Generation:}
    Integrate the spectral ODE system using high-precision numerical methods to generate time series of spectral coefficients for training.

    \item \textbf{Sequence Modeling:}
    Design the Transformer neural network that maps the past $S$ time steps of spectral coefficients and any auxiliary features to the next time step:
    \begin{equation}
        \{\hat{\mathbf{u}}_{t-\Delta t}, \dots, \hat{\mathbf{u}}_{t}\} \xrightarrow{\mathrm{NN}} \hat{\mathbf{u}}_{t+\Delta t}.
    \end{equation}

    \item \textbf{Network Architecture:}
    The Figure~\ref{fig:tran} shows the Fourier transformer architecture. The neural network is a Transformer based model that takes sequences of Fourier transformed data as input and predicts future Fourier modes.
    
    \item \textbf{Loss Function:}
    The loss function could be mean-squared error between predicted and true spectral coefficients at each prediction time step, or the residuals of the spectral ODE system:
    \begin{equation}
        \mathcal{L}_{\mathrm{phys}} = \left\| \frac{\hat{\mathbf{u}}_{t+\Delta t}^{\mathrm{pred}} - \hat{\mathbf{u}}_{t}^{\mathrm{input}}}{\Delta t} - \mathcal{F}\bigl(\hat{\mathbf{u}}_{t+\Delta t}^{\mathrm{pred}}, t+\Delta t\bigr) \right\|^2.
    \end{equation}

%    \item \textbf{Recursive Long-Term Prediction:}
%    During inference, recursively feed the network’s own predictions back as inputs to enable long-term forecasting.
\end{enumerate}

\subsection*{The Architecture of Neural Network}
The structure of the Fourier Spectral Transformer\cite{vaswani2017attention} in Figure~\ref{fig:tran} is introduced as follows:
%We adopt a Transformer based model for spectral coefficient prediction in Figure~\ref{fig:tran} as follows:

\paragraph{Input}
The input is a tensor of shape $B \times S \times N_1 \times N_2 \times \dots \times N_d \times D_{\text{input}}$, where $B$ is the batch size, $S$ is the sequence length, $N_1, N_2, \dots, N_d$ represent the number of spectral modes in each spatial direction for example, $N_1$ for the x-direction, $N_2$ for the y-direction, etc., and $D_{\text{input}}$ is the input dimension, typically corresponding to the features $[\Re(\hat{\mathbf{u}}_k), \Im(\hat{\mathbf{u}}_k), t]$, where $\Re(\hat{\mathbf{u}}_k)$ is the real part, $\Im(\hat{\mathbf{u}}_k)$ is the imaginary part, and $t$ is the time embedding.

\paragraph{Embedding}
Flatten the spatial and feature dimensions to obtain $B \times S \times N_1 \times N_2 \times \dots \times N_d \times D_{\text{input}}$, then apply a Dense layer to map into a latent space of dimension $d_{\text{model}}$:
\[
    \mathbf{N}_0 = \mathrm{Dense}\bigl(\mathbf{X}_{\mathrm{flat}}\bigr) \in \mathbb{R}^{B \times S \times d_{\text{model}}}.
\]

\paragraph{Stacked Self-Attention Layers}
Process the embedded sequence through $L$ layers, each consisting of multi-head self-attention (with $n_{\text{head}}$ heads) followed by layer normalization:
\[
    \mathbf{N}_{\ell} = \mathrm{LayerNorm}\bigl(\mathrm{SelfAttention}(\mathbf{N}_{\ell-1})\bigr), \quad \ell = 1, \dots, L.
\]

\paragraph{Decoding}
Extract the representation at the last time step: $\mathbf{N}_{\mathrm{out}} = \mathbf{N}_L[:, -1, :] \in \mathbb{R}^{B \times d_{\text{model}}}$, then use a final Dense layer to map back to $N_1 \times N_2 \times \dots \times N_d \times D_{\text{output}}$ and reshape to $B \times N_1 \times N_2 \times \dots \times N_d \times D_{\text{output}}$ for the next-step spectral coefficients.

This framework generalizes from PDEs to spectral ODEs and uses deep learning for efficient and physically consistent long term forecasting of complex dynamical systems.

\section{Results}\label{sec2}

\subsection{The 2D Navier-Stokes}
We consider the two-dimensional incompressible Navier--Stokes equations
\begin{eqnarray}
\frac{\partial{u}}{\partial t} +  u \frac{\partial u}{\partial x}+v\frac{\partial u}{\partial y}=\nu(\frac{\partial^2 u}{\partial x^2}+\frac{\partial^2 u}{\partial y^2}) - \frac{\partial p}{\partial x}\\
\frac{\partial{v}}{\partial t} +  u \frac{\partial v}{\partial x}+v\frac{\partial v}{\partial y}=\nu(\frac{\partial^2 v}{\partial x^2}+\frac{\partial^2 v}{\partial y^2}) -\frac{\partial p}{\partial y}
\end{eqnarray}
where $\mathbf{u}=(u,v)$ is the velocity field, $p$ the pressure, and $\nu$ the kinematic viscosity and the incompressible
\begin{eqnarray}
\frac{\partial u}{\partial x} + \frac{\partial v}{\partial y}= 0
\end{eqnarray}

The Fourier transform to the momentum equation yields
\[
\frac{\partial \widehat{\mathbf{u}}(\mathbf{k},t)}{\partial t}
= \widehat{\mathbf{N}}(\mathbf{k},t)
- \nu K^2\widehat{\mathbf{u}}(\mathbf{k},t)
- i\mathbf{k}\,\widehat{p}(\mathbf{k},t),
\]
where $\widehat{\mathbf{u}}=(\widehat{u},\widehat{v})$ and $\widehat{\mathbf{N}}$ is the Fourier transform of the nonlinear term.

\begin{eqnarray}
K_1\hat u + K_2\hat v = 0
\end{eqnarray}

\paragraph{The Numerical Fourier Spectral Simulation}

\begin{figure}
  \centering
  \foreach \img/\lettr in {
    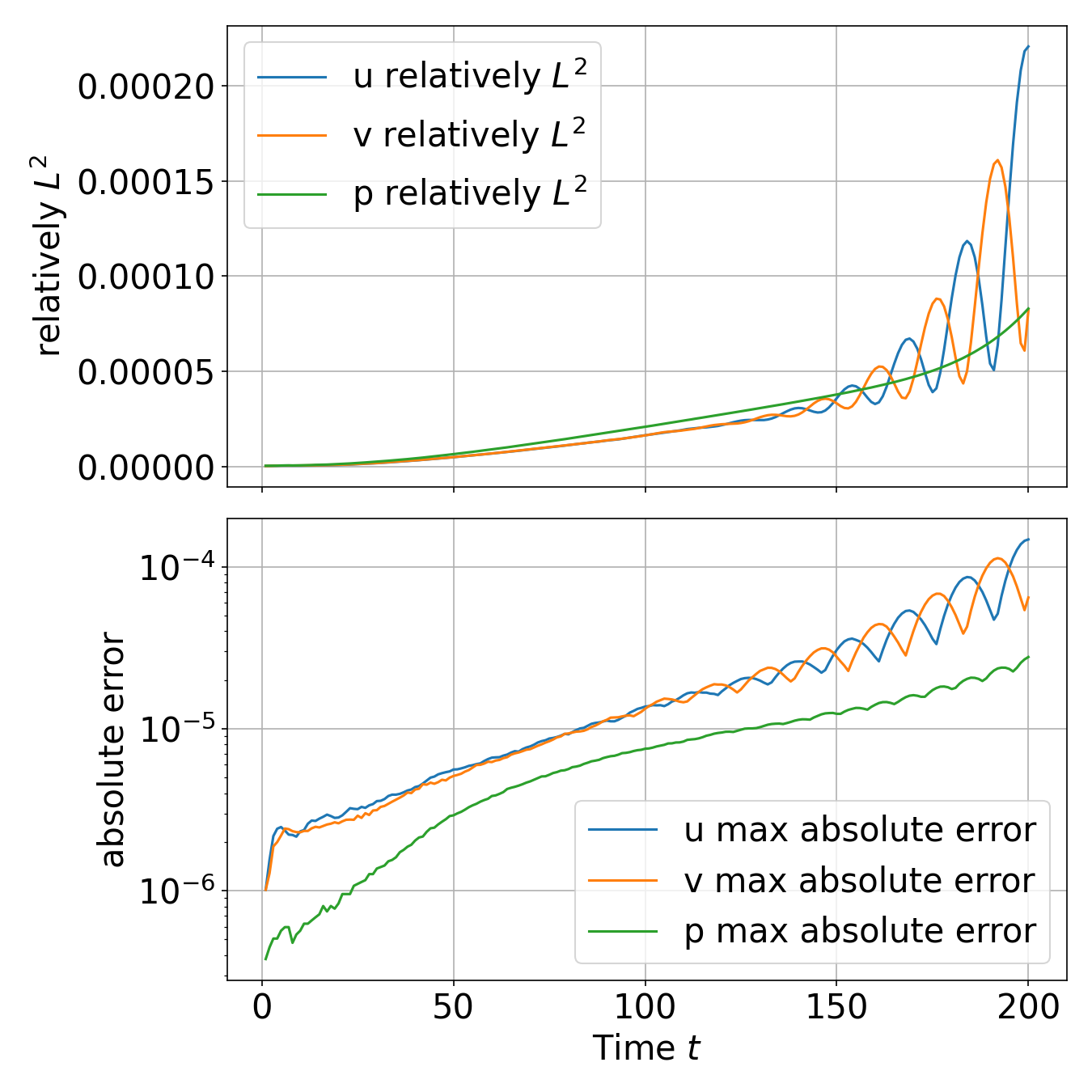/a,
    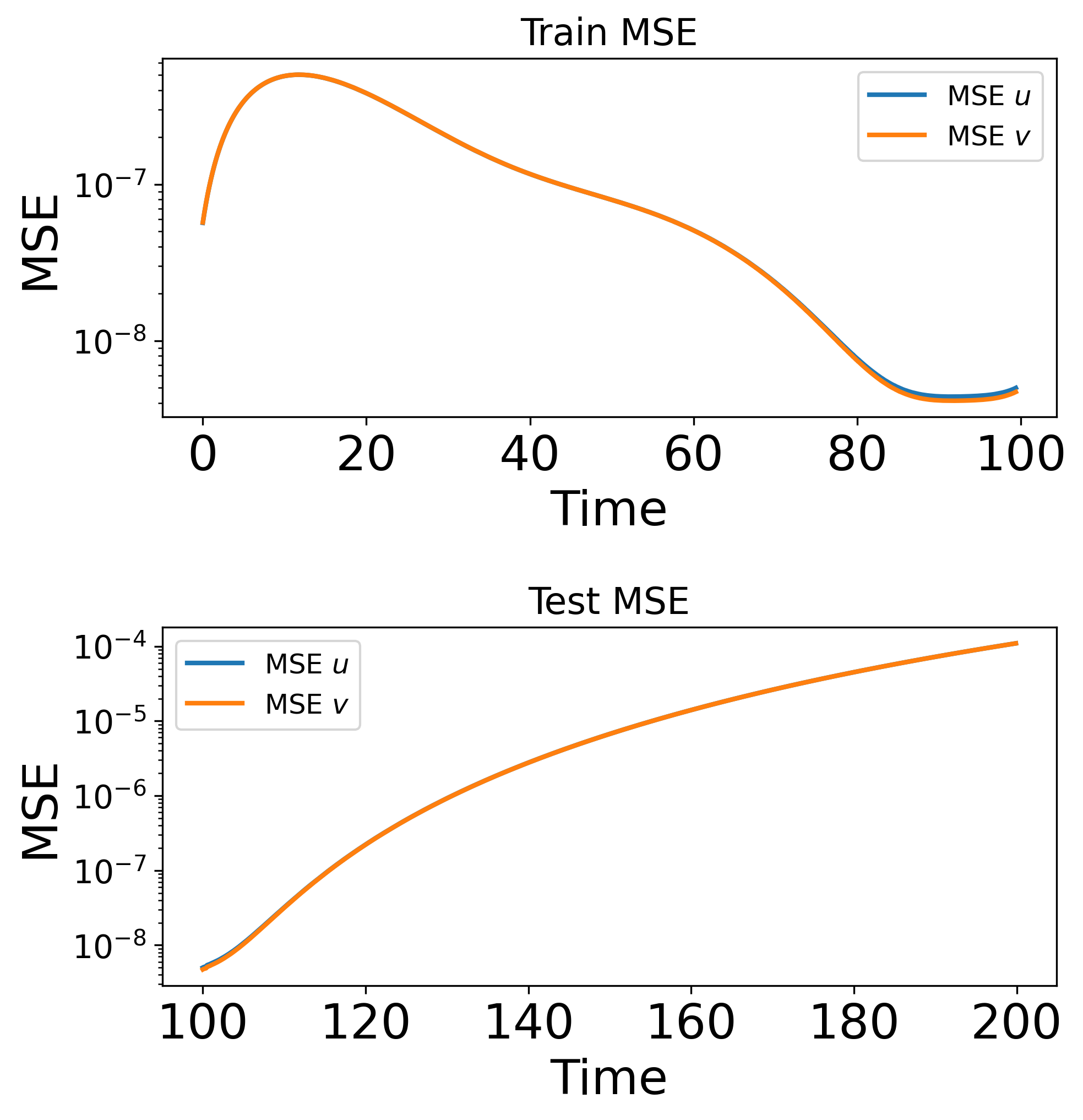/b
  }{
    \begin{tikzpicture}[baseline=(X.north)]
      \node (X) {\includegraphics[width=0.45\textwidth]{\img}};
      \node[anchor=north west,inner sep=2pt,font=\small\bfseries]
        at (X.north west) {\lettr};
    \end{tikzpicture}\quad
  }\par\vspace{0.5em}

  \caption{\textbf{The analytical and numerical error. The numerical and transformer error.}
    \textbf{a}, The relative $L^2$ error and maximum absolute error of the spectral 2D Navier Stokes solver;  
    \textbf{b}, The MSE between Transformer and numerical method for train set from 0 to 100 and test set from 100 to 200.
  }
  \label{fig:numer}
\end{figure}

We consider the two-dimensional incompressible Navier--Stokes equations on a periodic domain $[ -\pi, \pi ]^2$. We validate our Fourier spectral solver against the analytical Taylor–Green vortex solution:
\begin{eqnarray}\label{TV}
 u(x,y,t)=\sin x\,\cos y\,e^{-2\nu t}, \ \ 
 v(x,y,t)=-\cos x\,\sin y\,e^{-2\nu t} \nonumber\\
 p(x,y,t)=-\tfrac14\bigl(\cos2x+\cos2y\bigr)e^{-4\nu t}.
\end{eqnarray}
The Figure~\ref{fig:numer}a shows the relative $L^2$ errors and maximum absolute errors are around $10^{-5} $ for $u$, $v$, and $p$ over $t\le200.0$. The solver retains spectral accuracy in space and fourth‐order accuracy in time. Time integration is carried out using a fourth-order Runge Kutta scheme with time step $0.1$ and viscosity $\nu = 10^{-3}$. The simulation is run for $2000$ time steps to generate reference ground truth data.
%\begin{align}
%    \frac{\partial \mathbf{u}}{\partial t} + (\mathbf{u} \cdot \nabla) \mathbf{u} &= -\nabla p + \nu \nabla^2 \mathbf{u}, \\
%    \nabla \cdot \mathbf{u} &= 0,
%\end{align}
%where $\mathbf{u} = (u, v)$ is the velocity field, $p$ is the pressure, and $\nu$ is the kinematic viscosity. 

%The initial condition is given by
%\begin{align}
%    u(x, y, 0) &= \sin(x)\cos(y), \\
%    v(x, y, 0) &= -\cos(x)\sin(y).
%\end{align}

%Representative snapshots of $u$, $v$, and $p$ at $t=0$ and $t=200.0$ are shown in Figure~\ref{fig:snapshots}.  As viscosity acts, the vortex strength and pressure amplitude decay exponentially. 

%We discretize the domain with $N=64$ grid points in each direction. The equations are solved in spectral space, with nonlinear terms computed in physical space and then transformed to Fourier space. 
 
%The numerical solver is very accurate which has been compared with analytical solution in Figure~\ref{fig:numer}. 

%\begin{figure}
%    \centering
%     \includegraphics[width=0.45\textwidth]{numerical.png}
%     \includegraphics[width=0.45\textwidth]{mse_curves.png}
%    \caption{}
%    \label{fig:numer}
%\end{figure}

%\paragraph{Data Preparation}
%At each time step, the Fourier coefficients for $u$ and $v$ are stored. For training, we use a sliding window of length $5$: each input sample consists of $5$ consecutive time steps, where each step includes the real and imaginary parts of $\hat{u}$ and $\hat{v}$, along with a normalized time channel. The target output is the spectral coefficients at the next time step. Formally, each input sample has shape $[5, N, N, 5]$, where the last dimension contains $[\mathrm{Re}(\hat{u}), \mathrm{Im}(\hat{u}), \mathrm{Re}(\hat{v}), \mathrm{Im}(\hat{v}), t_{\mathrm{norm}}]$.

%\paragraph{The Fourier Spectral Transformer Neural Network Architecture}

%We construct a physics informed neural network based on a Transformer encoder, referred to as \texttt{Spectral2DTransformer}. The model processes inputs of shape $[\text{batch}, 5, N, N, 5]$ and outputs the spectral coefficients $[\text{batch}, N, N, 4]$ for the next time step. The network consists of two Transformer encoder layers with four attention heads and a model dimension of $128$. Only the last frame in the input sequence is used for final prediction.

\paragraph{The Loss Function of Transformer}
We've tried the MSE loss between the Transformer outputs and numerical results, and the ODE residual loss plus the initial condition loss. They have similar results in which the MSE loss is slightly better than the ODE residual plus the initial condition loss. 

\paragraph{The Mean Squared Error}

The mean-squared error between the predicted and true spectral coefficients is computed at each time step:
\begin{equation}
    \mathrm{MSE}(t) = \frac{1}{N^2} \sum_{i,j} \left| \hat{u}_{\mathrm{pred}}^{i,j}(t) - \hat{u}_{\mathrm{true}}^{i,j}(t) \right|^2 + \left| \hat{v}_{\mathrm{pred}}^{i,j}(t) - \hat{v}_{\mathrm{true}}^{i,j}(t) \right|^2.
\end{equation}
Figure~\ref{fig:numer}b shows the evolution of MSE over time.

%\begin{figure}
%    \centering
%     \includegraphics[width=0.45\textwidth]{train_mse.pdf}
%     \includegraphics[width=0.45\textwidth]{pred_mse.pdf}
%    \caption{Temporal evolution of mean-squared error (MSE) between PINN predictions and spectral ground truth. The red dashed line marks the end of the training interval.}
%    \label{fig:mse_time}
%\end{figure}

\paragraph{Training And Prediction Set Comparison}
The first set of results compares the predictions of the neural network with the spectral method on the training set, spanning the first 1000 time steps, for $t$ from $0$ to $100$. The neural network's predictions are visually compared with the spectral method's true solution at selected time intervals. 

\begin{figure}
    \centering
    \includegraphics[width=\textwidth]{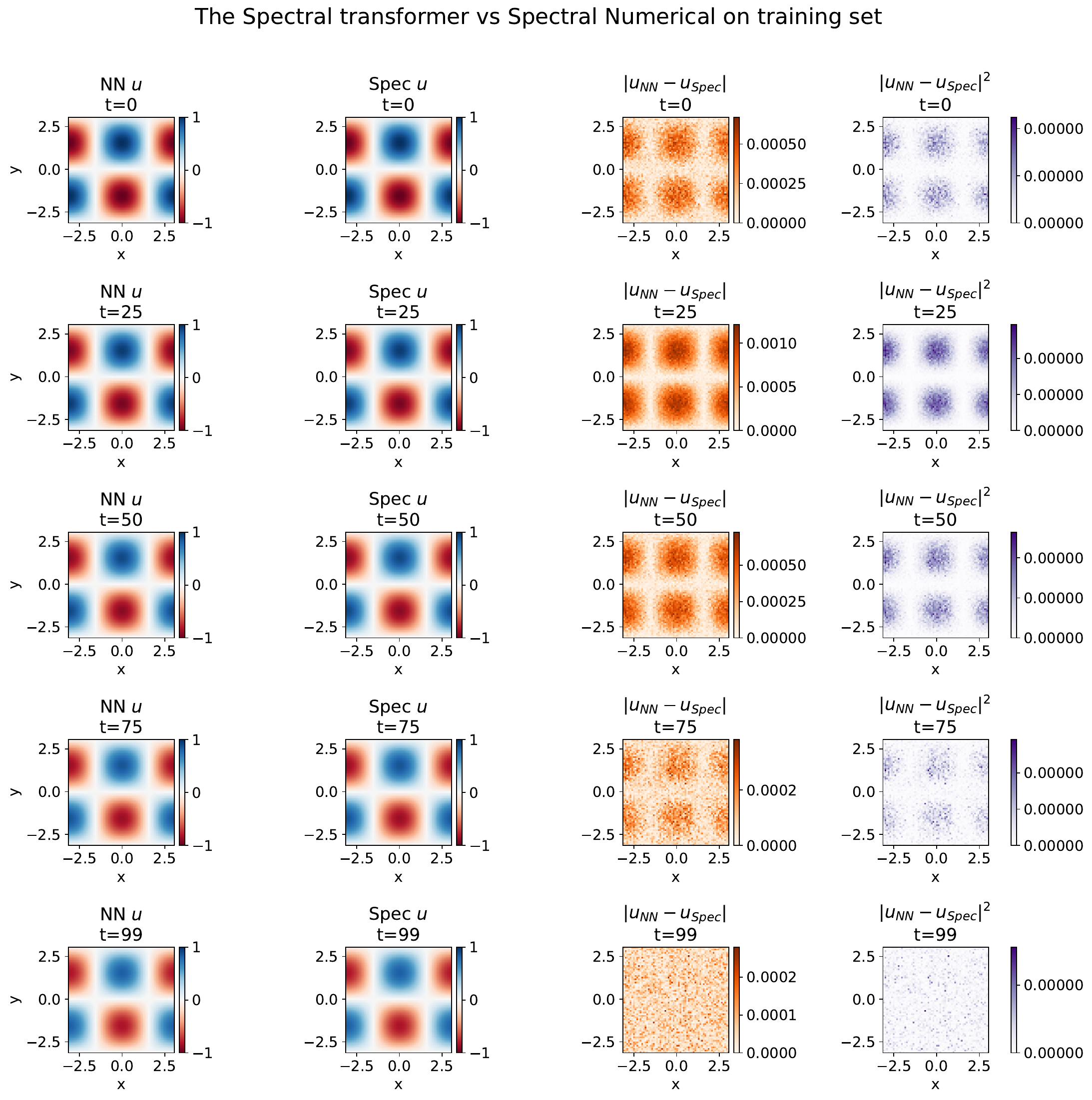}
    \caption{The Fourier spectral transformer results of $u$, the numerical results of $u$,  the absolute error and the mean squared error. The training set time is from 0 to 99.}
    \label{fig:u}
\end{figure}

\begin{figure}
    \centering
    \includegraphics[width=\textwidth]{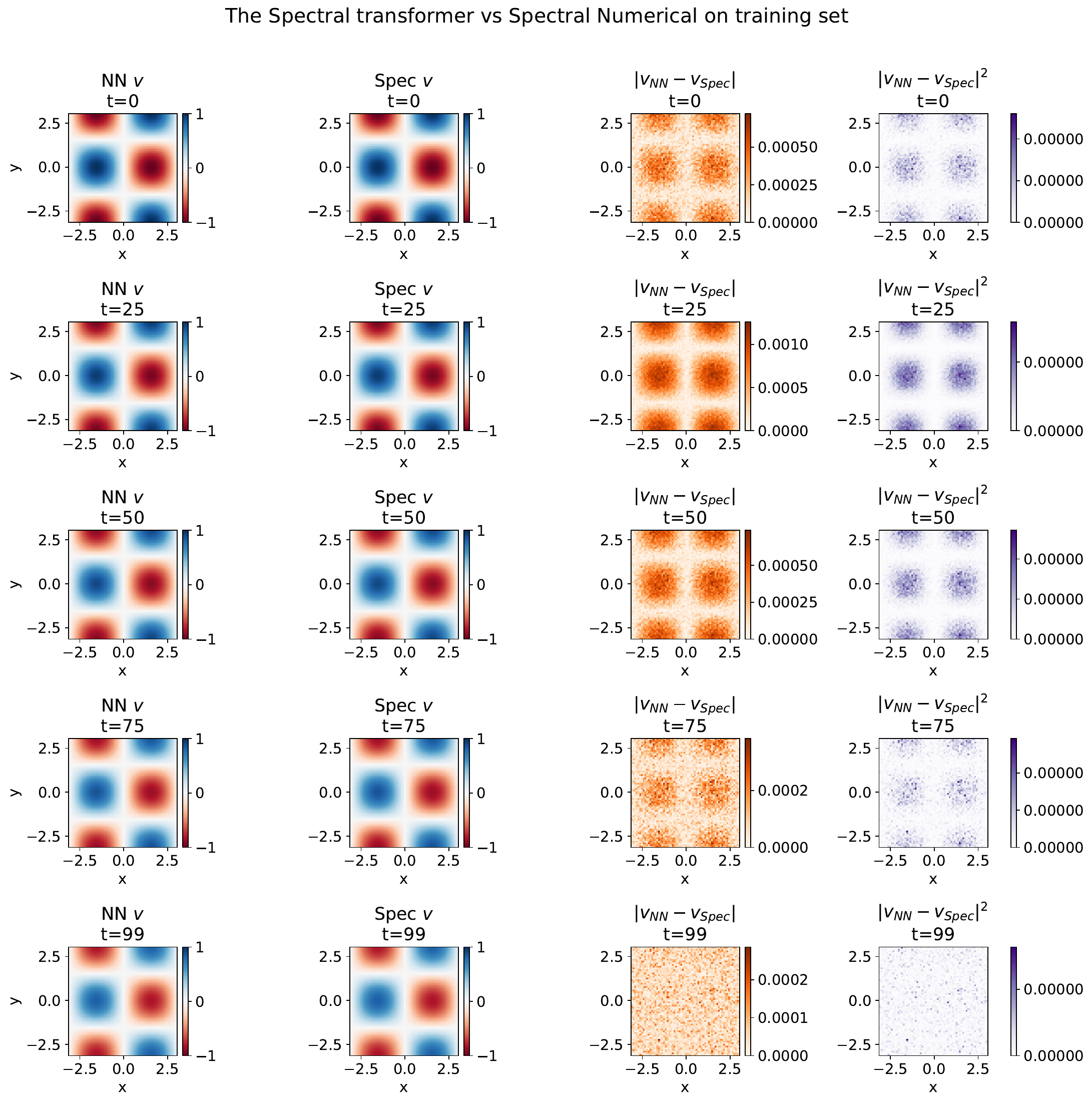}
    \caption{The Fourier spectral transformer results of $v$, the numerical results of $v$,  the absolute error and the mean squared error. The training set time is from 0 to 99.}
    \label{fig:v}
\end{figure}

The Figures~\ref{fig:u} and~\ref{fig:v} show the error distributions for the predicted velocity components $u$ and $v$, highlighting the accuracy of the NN model compared to the spectral method. The NN model is able to capture the dynamics of the flow with minimal error across the training set.

\begin{figure}
    \centering
    \includegraphics[width=\textwidth]{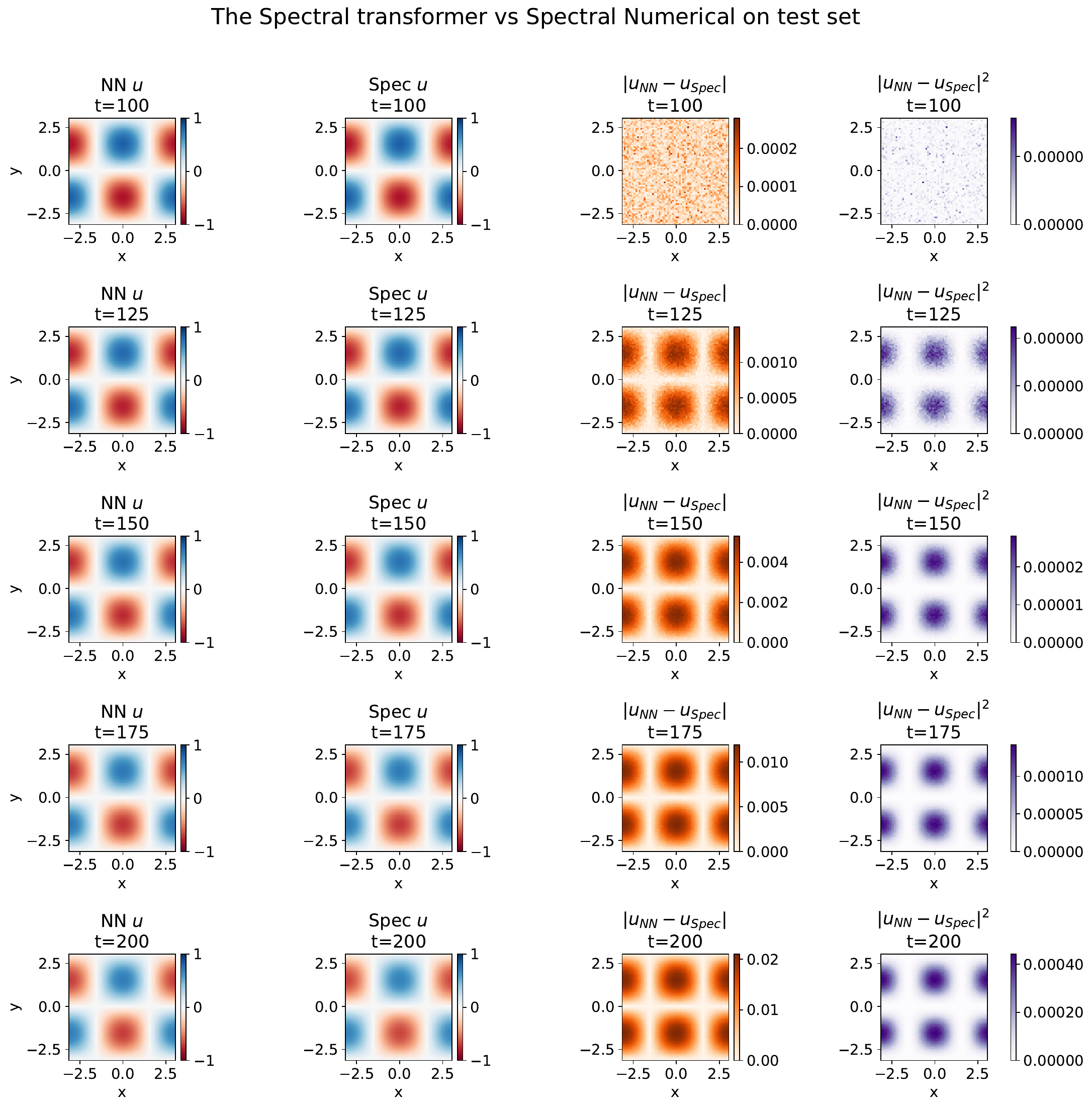}
    \caption{The Fourier spectral transformer results of $u$, the numerical results of $u$,  the absolute error and the mean squared error. The test set time is from 100 to 200.}
    \label{fig:ut}
\end{figure}

\begin{figure}
    \centering
    \includegraphics[width=\textwidth]{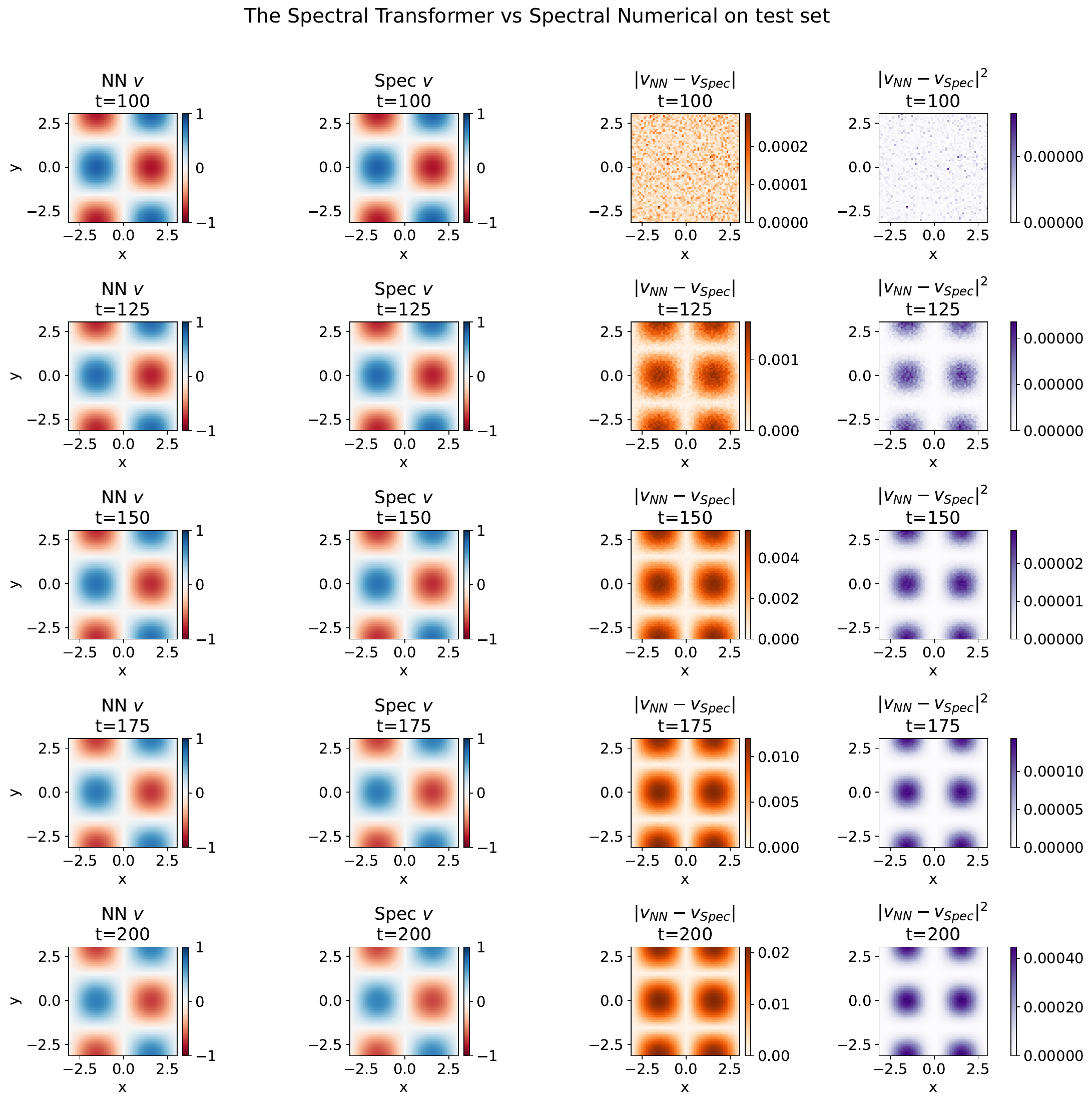}
    \caption{The Fourier spectral transformer results of $v$, the numerical results of $v$,  the absolute error and the mean squared error. The test set time is from 100 to 200.}
    \label{fig:vt}
\end{figure}

We next evaluate the neural network's performance on the prediction set in Figures~\ref{fig:ut} and~\ref{fig:vt}, which spans the next 1000 time steps for $t$ from $100$ to $200$. The neural network is tasked with forecasting future velocity fields given initial conditions at the start of the prediction period.

The Figures illustrate the error distributions for the predicted fields compared to the true fields from the spectral method. The transformer captures the main flow features and maintains reasonable accuracy over long term extrapolation. 


This shows the spectral Fourier transformer has the potential to predict long time after training, which is better than traditional numerical methods. The neural network is capable of producing highly accurate predictions even when trained on relatively limited data. The model's ability to generalize from temporal sequences allows it to make robust predictions of future velocity fields with minimal computational cost, a significant advantage over traditional methods. This capability is particularly valuable for forecasting fluid behaviors in scenarios where real-time predictions are needed. As demonstrated in the results, the neural network is good at capturing the dynamics of the flow across both the training and prediction intervals, further proving its potential as a powerful tool for solving complex fluid dynamics problems.

\subsection{Burgers' equation}
The Burgers equation 
\begin{equation}
	\frac{\partial u}{\partial t}+u\frac{\partial u}{\partial x} = \nu\frac{\partial ^2 u}{\partial x^2},  \ \ \ \ \ -\pi  \leq  x \leq \pi, \ \ t > 0
\end{equation}
with boundary conditions
\begin{eqnarray}
	u(\pi,t)=u(-\pi,t)=0
\end{eqnarray}
and initial condition
\begin{eqnarray}
	u(x,0)=-\sin(x)
\end{eqnarray}
where $\nu=\pi\nu=10^-2$. 
The solution could be approximated as a truncated Fourier series
\begin{equation}
	u(x,t)=\sum_{k=-N}^{N} \hat u_k(t)e^{ikx}.
\end{equation}

\subsubsection*{The Fourier Spectral Transformer Results}

\begin{figure}
        \includegraphics[width=0.5\linewidth]{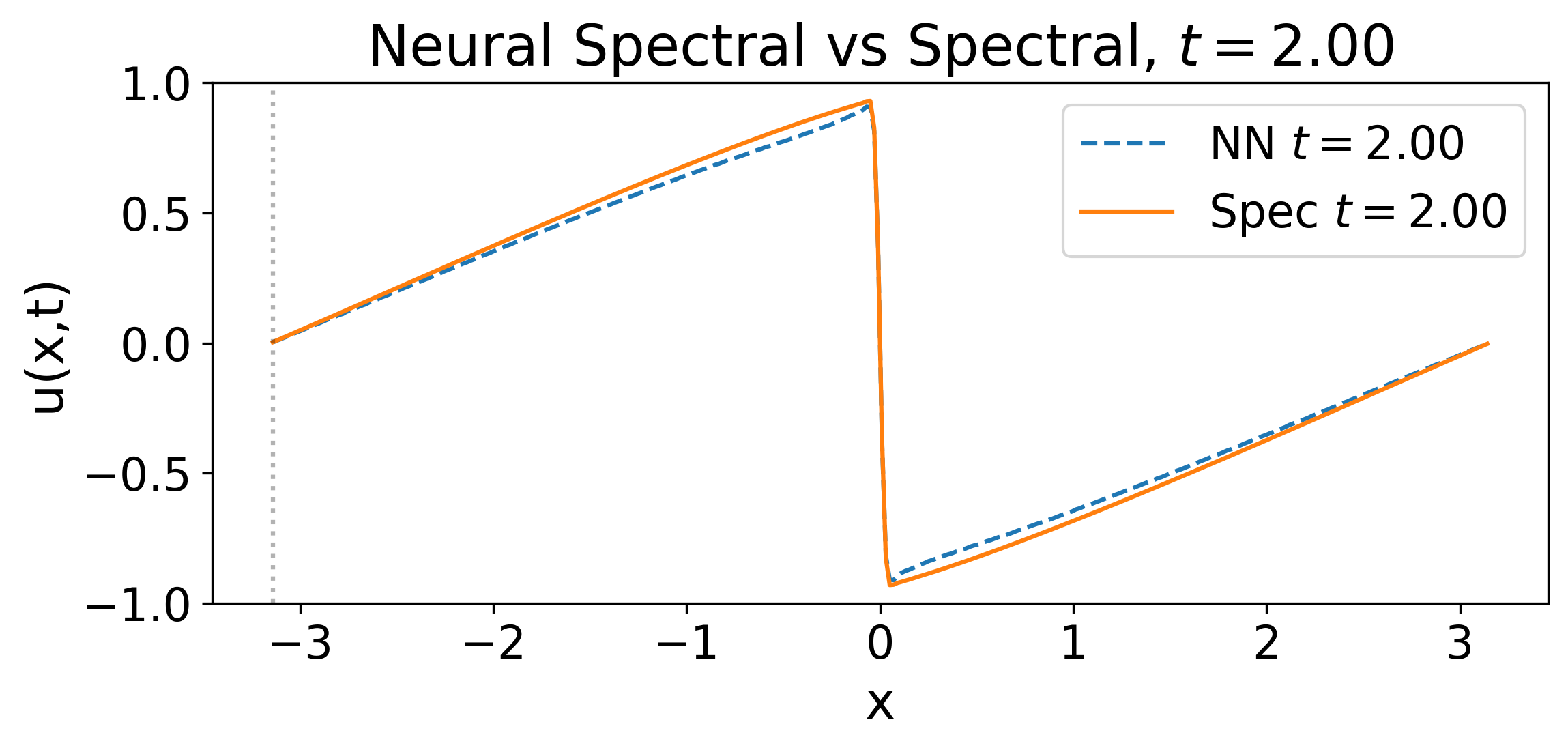}
        \includegraphics[width=0.5\linewidth]{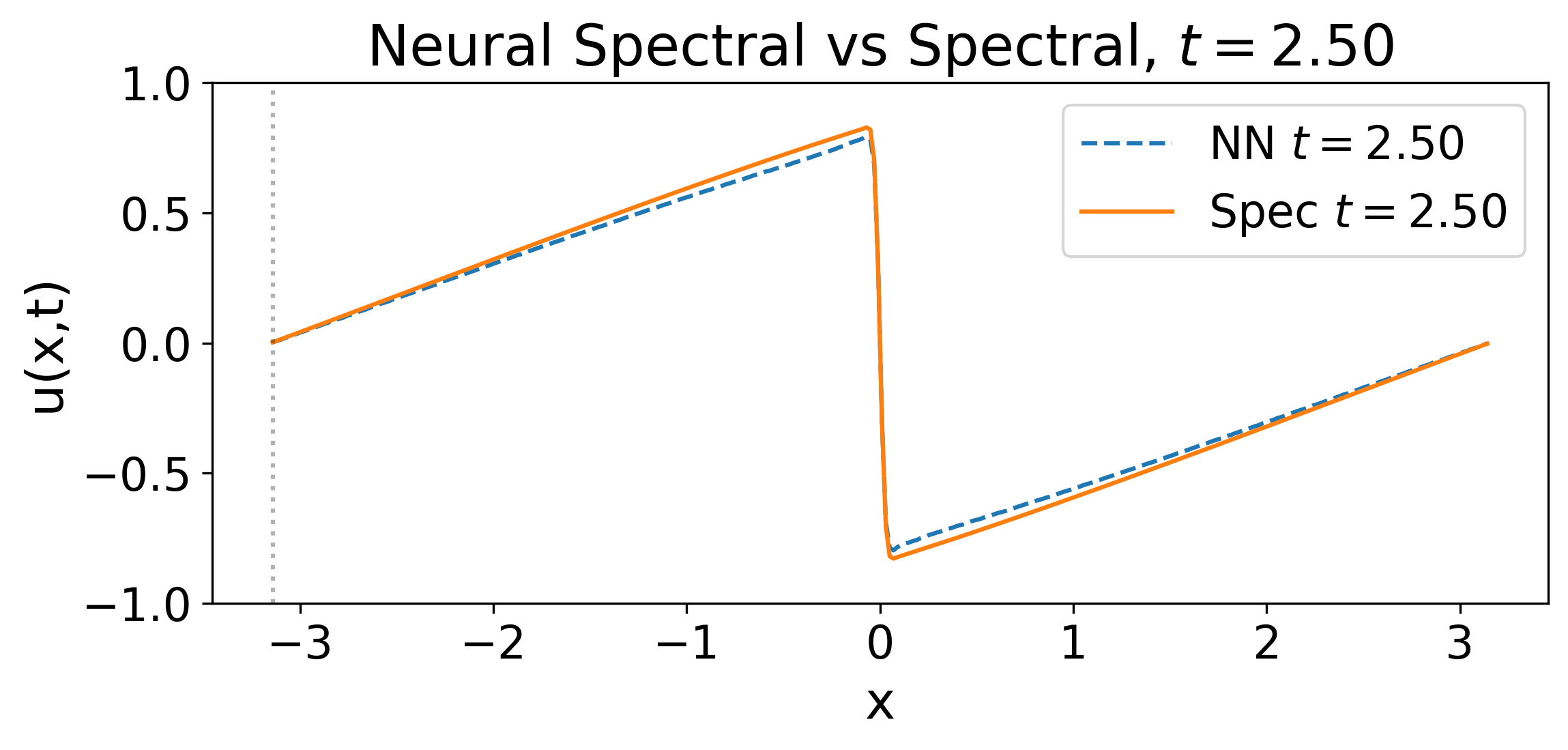}
        \includegraphics[width=0.5\linewidth]{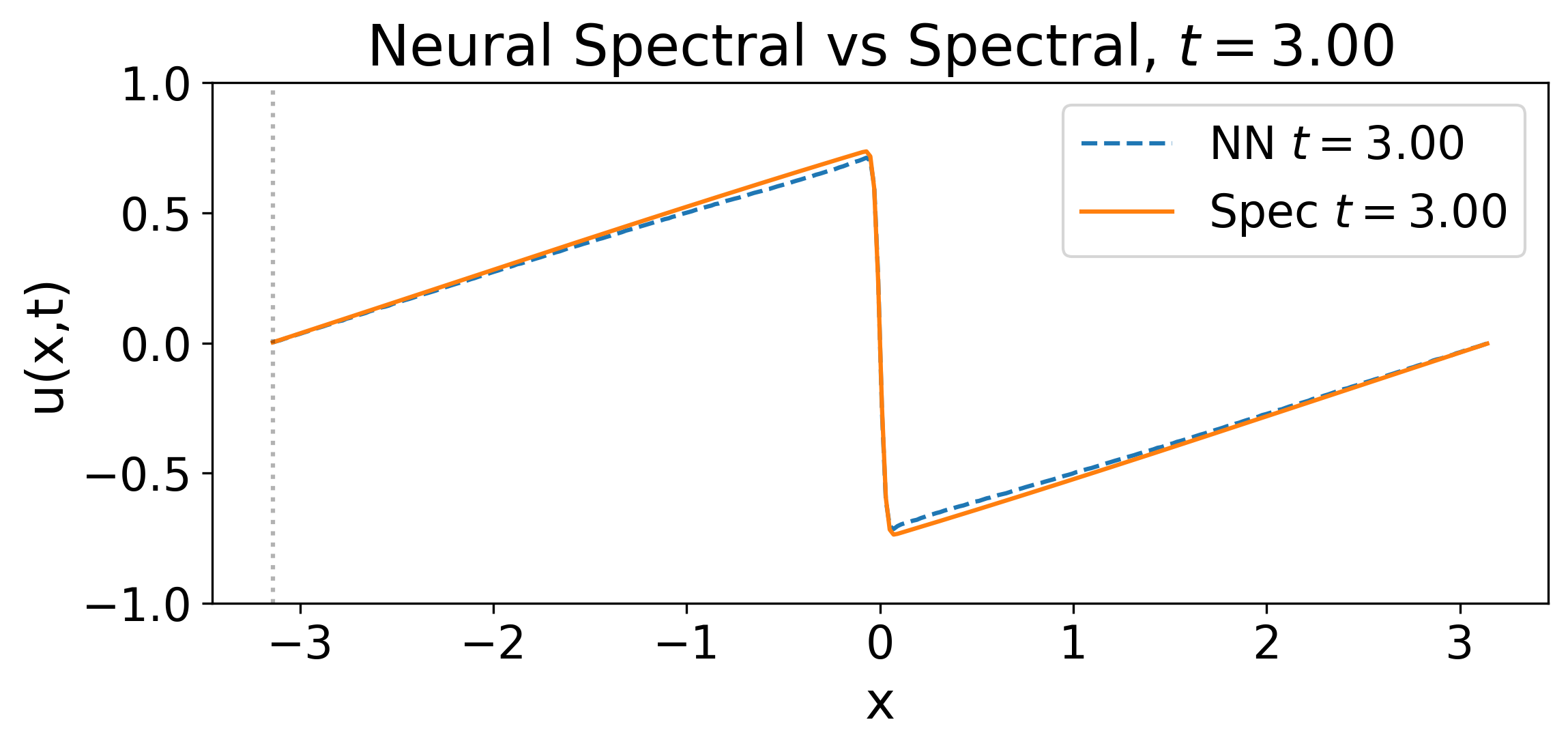}
        \includegraphics[width=0.5\linewidth]{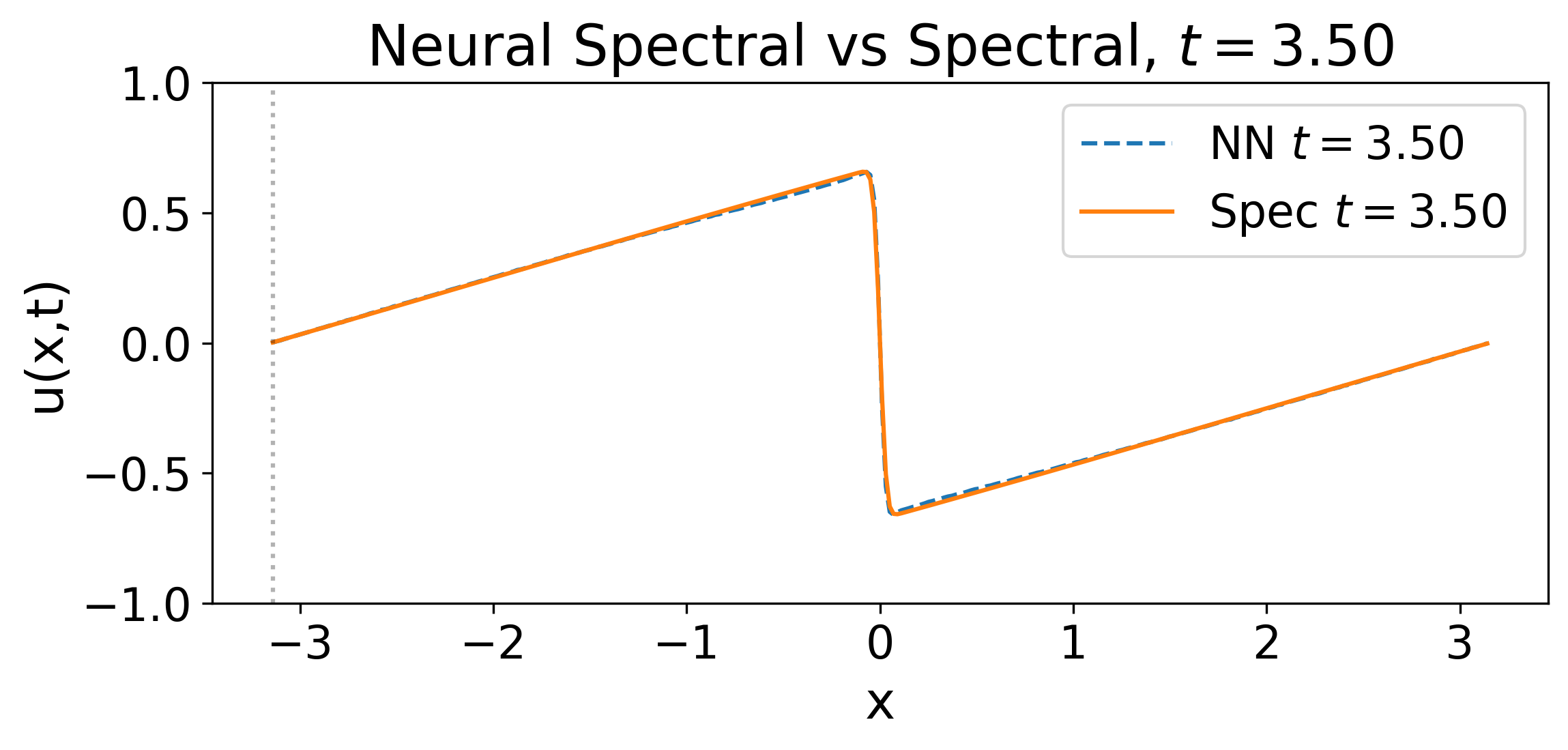}
        \includegraphics[width=0.5\linewidth]{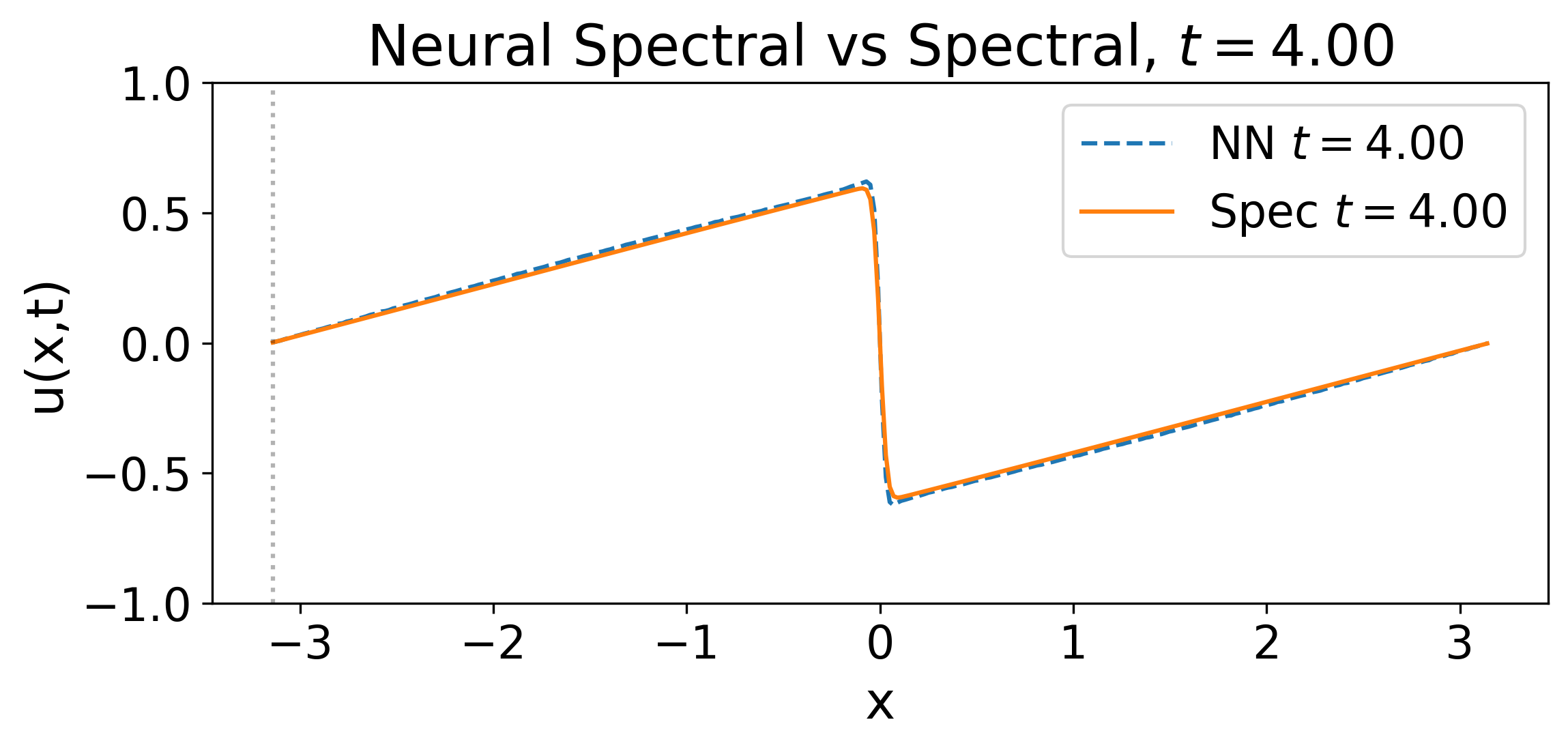}
        \includegraphics[width=0.5\linewidth]{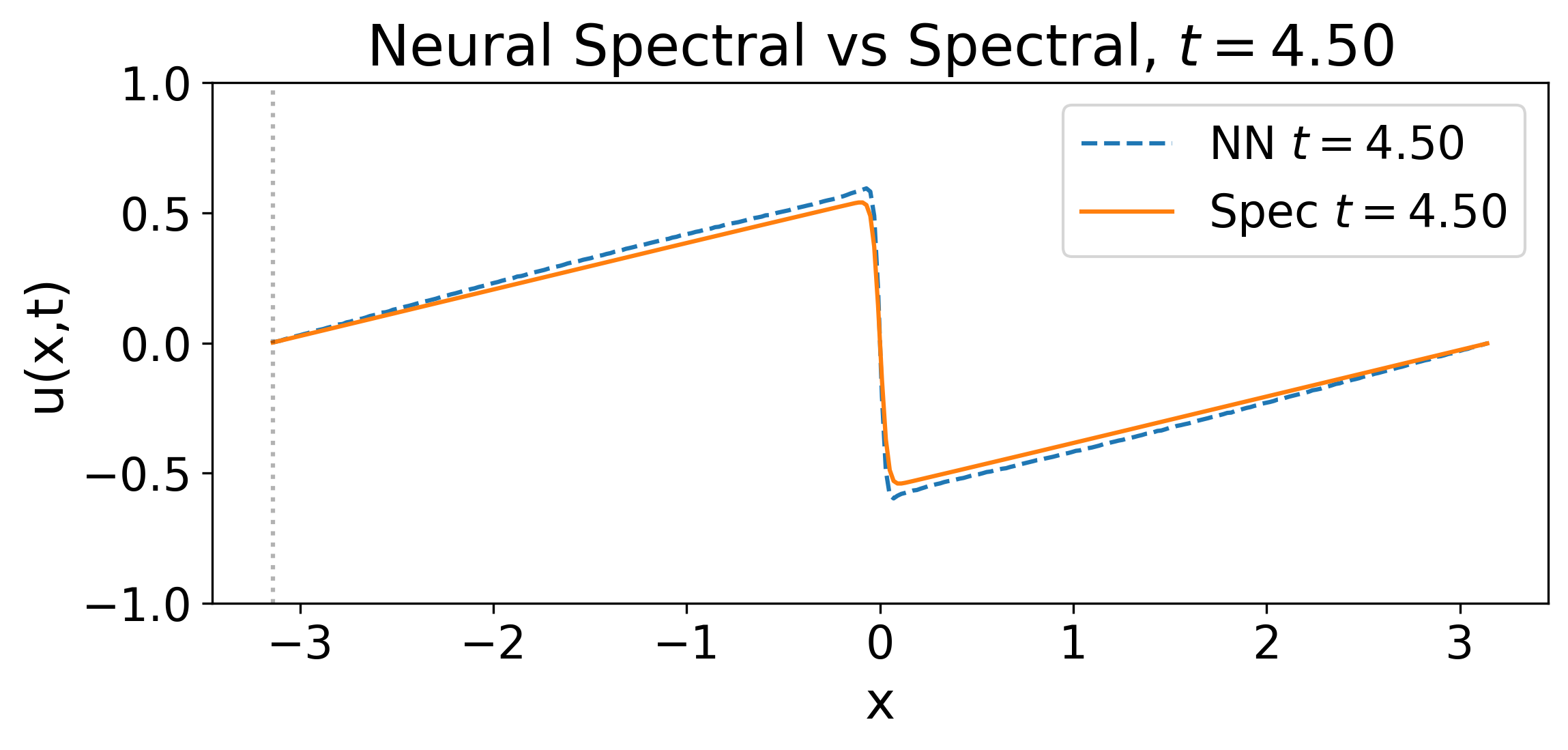}
        \includegraphics[width=0.5\linewidth]{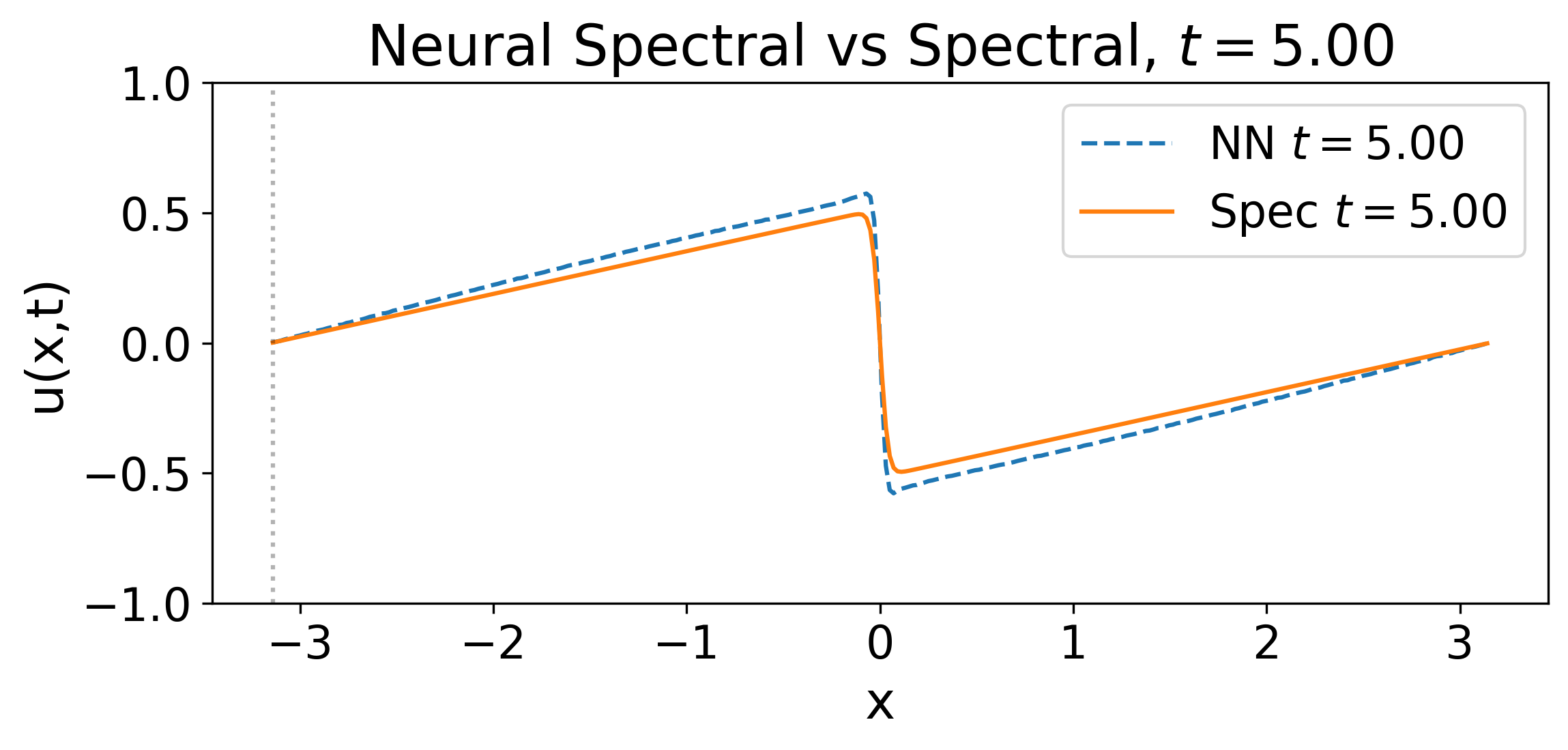}
        \includegraphics[width=0.5\linewidth]{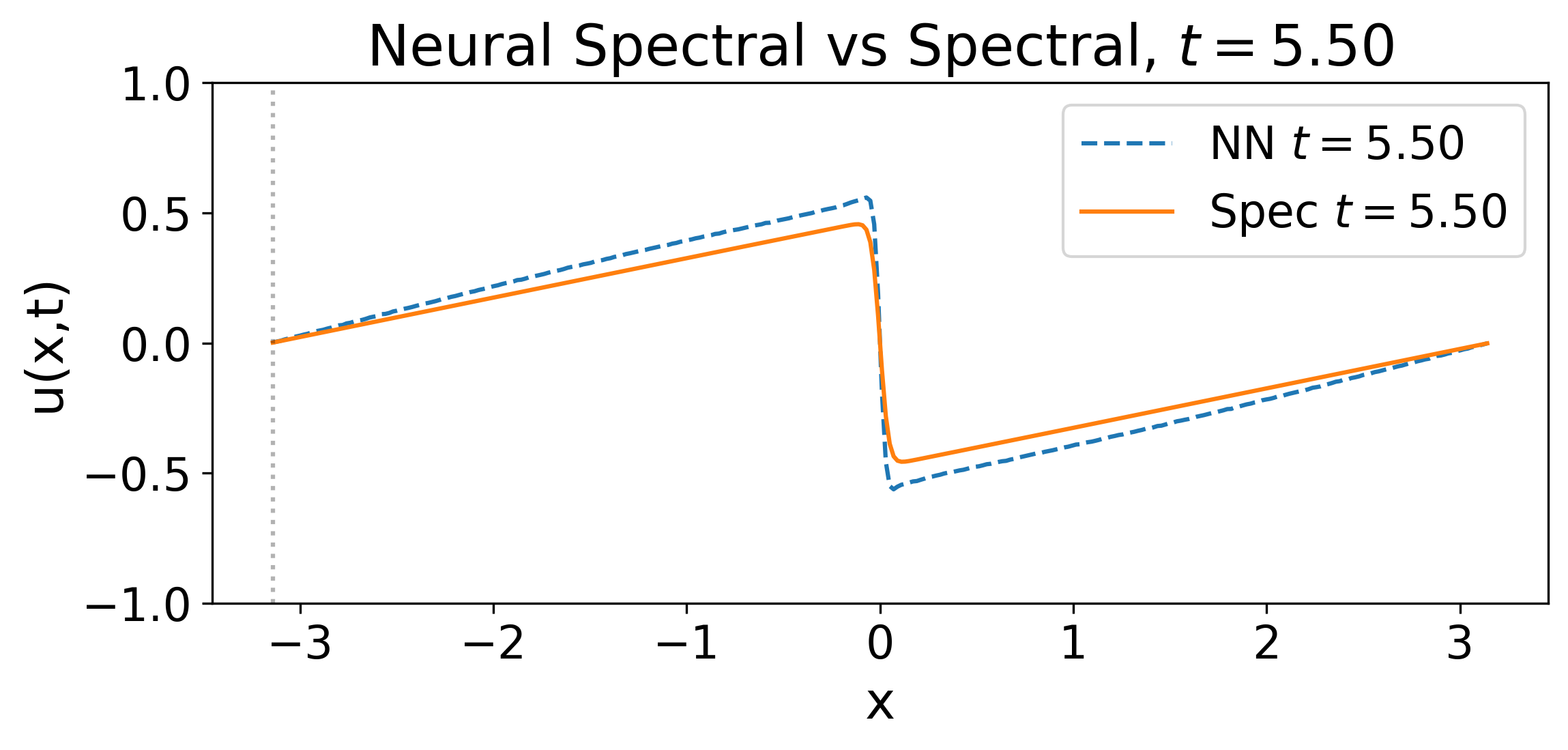}
    \caption{Neural spectral vs spectral solutions at selected time slices. The training set is $[0,3]$, and the test set is $[3, 5.5]$.}
    \label{fig:burgers}
\end{figure}

In this experiment we compare the long term prediction capabilities of the Fourier spectral transformer with the classical spectral method for solving the 1D Burgers' equation. Training data is generated via the Fourier spectral method for the interval \( t \in [0, 3] \). Each training sequence consists of both the real and imaginary parts of the spectral coefficients, as well as a normalized time embedding, which are fed as multi-channel input to a Transformer neural network implemented with JAX and Flax. 
After training, the network is autoregressively rolled out to predict the system dynamics well beyond the training window in Figure~\ref{fig:burgers}.

%The neural network is trained using a loss function defined as the ODE residual of the Burgers' equation in spectral space. This ensures that the model's prediction of spectral coefficients obeys the physical law without the need for explicit ground truth labels. 
%Comparative visualizations show that the physics-informed neural network closely tracks the traditional spectral solution at multiple representative time slices. 

\begin{figure}
    \centering
    \includegraphics[width=0.7\linewidth]{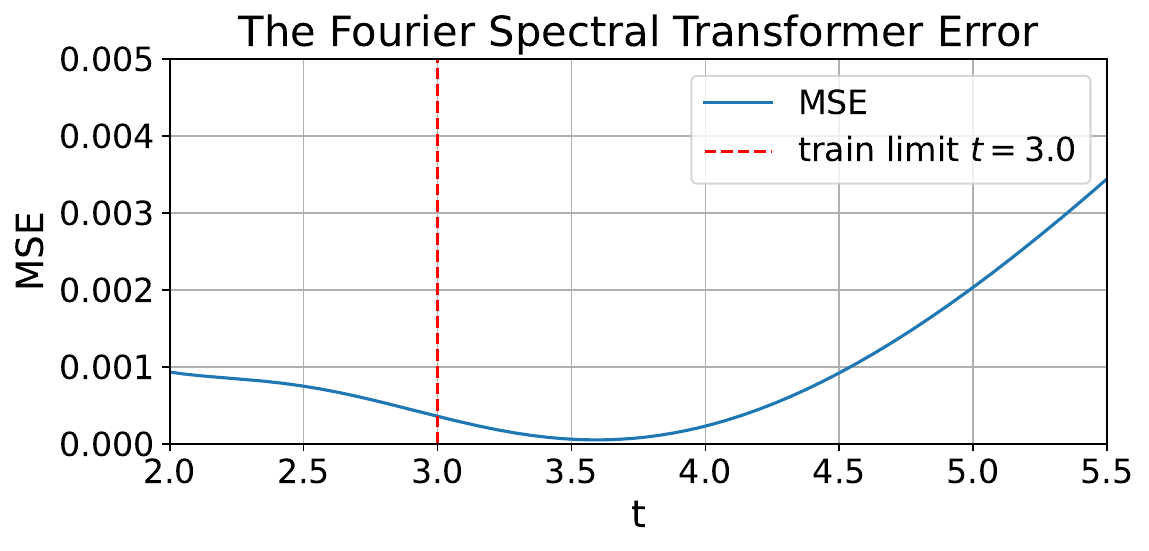}
    \caption{The Mean squared error between neural spectral and spectral solutions for $t \in [2,5.5]$. The dashed line indicates the end of the training interval $t=3.0$.}
    \label{fig:mse_curve}
\end{figure}

The evolution of mean squared error between the two methods is plotted over in Figure~\ref{fig:mse_curve}. The results demonstrate that the physics-constrained neural network exhibits strong generalization and physical consistency, offering a promising approach for modeling and forecasting nonlinear dynamical systems beyond the training domain.

The total loss used for training is the ODE residual loss and initial condition loss.
The model's performance is evaluated by comparing the predicted flow fields with ground truth data, and the temporal MSE is computed to assess the prediction accuracy over time. The results show that the neural network captures the main flow features and maintains reasonable accuracy for long term extrapolation.

\section{Conclusion}
%We have introduced a Fourier Spectral Transformer network that unifies spectral numerical methods with modern neural network architectures for modeling and forecasting solutions to PDEs. Through extensive experiments on both the 2D Navier-Stokes and 1D Burgers’ equations, we show that our method not only achieves spectral-level accuracy during the training interval, but also generalizes well for long-term predictions beyond the training window. The model is able to efficiently capture the dynamics of nonlinear systems, offering a significant reduction in computational cost compared to traditional numerical solvers. Our findings suggest that physics-informed neural operators, when combined with classical spectral techniques, provide a powerful framework for solving complex fluid dynamics problems. Future work will focus on extending this approach to more challenging systems and exploring theoretical guarantees for stability and generalization.

We have developed a unified Fourier Spectral Transformer network that combines the mathematical rigor of spectral methods with the flexibility of Transformer neural networks for solving and forecasting partial differential equations. The approach transforms the original PDEs into spectral ODEs and uses a sequence modeling framework to predict the future evolution of spectral coefficients. Through comprehensive experiments on the two-dimensional incompressible Navier-Stokes equations and the one-dimensional Burgers’ equation, we have demonstrated that the proposed model achieves high accuracy both during training and in long term prediction better than other machine learning methods and traditional numerical methods in forecasting capability. The neural network is able to generalize well from limited data and efficiently capture the essential dynamics of complex systems with significantly reduced computational cost. These results suggest that the integration of Fourier spectral methods and transformer is a promising direction for real time simulation and control of nonlinear dynamical systems. Future research will focus on extending this framework to higher dimensional problems, more complex PDEs, and providing rigorous theoretical analysis of stability and convergence.
\bibliography{Untitled}
\end{document}